\documentclass{article}
\usepackage{ijcai17}

\usepackage{times}

\usepackage{macros}

\usepackage{latexsym}

\title{Thresholding Bandits with Augmented UCB}

\author{\Large Subhojyoti Mukherjee${}^1$,Naveen Kolar Purushothama${}^2$,Nandan
Sudarsanam${}^3$,Balaraman Ravindran${}^4$ \\
\Large ${}^{1,4}$Department of Computer Science \& Engineering, Indian Institute of
Technology Madras\\ 
\Large ${}^2$Department of Electrical Engineering, Indian Institute of
Technology Madras\\
\Large ${}^3$Department of Management Studies, Indian Institute of
Technology Madras
\\
\Large ${}^{1}$subho@cse.iitm.ac.in, ${}^{2}$naveenkp@iitm.ac.in, ${}^{3}$nandan@iitm.ac.in, ${}^{4}$ravi@cse.iitm.ac.in
}

\begin{document}
\maketitle

\vspace*{2mm}
\begin{abstract}
In this paper we propose the Augmented-UCB (AugUCB) algorithm for a fixed-budget version of the thresholding bandit problem (TBP), where the objective is to identify a set of arms whose quality is above a threshold. A key feature of AugUCB is that it uses both mean and variance estimates to eliminate arms that have been sufficiently explored; to the best of our knowledge this is the first algorithm to employ such an approach for the considered TBP.  Theoretically, we obtain an upper bound on the loss (probability of mis-classification) incurred by AugUCB. Although UCBEV in literature provides a better guarantee, it is important to emphasize that UCBEV has access to problem complexity (whose computation requires arms' mean and variances), and hence is not realistic in practice; this is in contrast to AugUCB whose implementation does not require any such complexity inputs. We conduct extensive simulation experiments to validate the performance of AugUCB. Through our simulation work, we establish that AugUCB, owing to its utilization of variance estimates, performs significantly better than the state-of-the-art APT, CSAR and other non variance-based algorithms.
\end{abstract}


\section{Introduction}
\label{intro}
Stochastic multi-armed bandit (MAB) problems are instances of the classic sequential decision-making scenario; specifically an MAB problem comprises of a learner and a collection of actions (or arms), denoted $\mathcal{A}$. In each trial the learner plays (or pulls) an arm $i\in\mathcal{A}$ which yields independent and identically distributed (i.i.d.) reward samples from a distribution (corresponding to arm $i$), whose expectation is denoted by $r_i$. \blfootnote{Accepted at the Proceedings of the Twenty-Sixth International Joint Conference on Artificial Intelligence (IJCAI-17). Copyright 2017 by the author(s).} The learner's objective is to identify an arm corresponding to the maximum expected reward, denoted $r^{*}$. Thus, at each time-step the learner is faced with the \emph{exploration vs.\ exploitation dilemma}, where it can pull an arm which has yielded the highest mean reward (denoted $\hat{r}_{i}$) thus far (\emph{exploitation}) or continue to explore other arms with the prospect of finding a better arm whose performance has not been observed sufficiently (\emph{exploration}).

Pure-exploration MAB problems are unlike their traditional (exploration vs.\ exploitation)  counterparts where the  objective is to minimize the cumulative regret (which is the total loss incurred by the learner for not playing the optimal arm throughout the time horizon $T$). Instead, in pure-exploration problems a learning algorithm, until time $T$, can invest entirely on exploring the arms without being concerned about the loss incurred while exploring; the objective is to minimize the probability that the arm recommended at time $T$ is not the best arm.  In this paper, we further consider a combinatorial version of the pure-exploration MAB, called the thresholding bandit problem (TBP).  Here, the learning algorithm is provided with a threshold $\tau$, and the objective, after exploring for $T$ rounds, is to  output all arms $i$ whose $r_{i}$ is above $\tau$. 
It is important to emphasize that the \emph{thresholding} bandit problem is different from the \emph{threshold} bandit setup studied in \cite{abernethy2016threshold}, where the learner receives an unit reward whenever the value of an observation is above a threshold.

Formally, the problem we consider is the following. First, we define the set $S_{\tau}=\lbrace i\in \mathcal{A}: r_{i}\geq \tau \rbrace$. Note that, $S_\tau$ is the set of all arms whose reward mean is greater than $\tau$. Let 
$S_\tau^c$ 
 denote the complement of $S_\tau$, i.e.,  $S_{\tau}^{c}=\lbrace i\in \mathcal{A}: r_{i} < \tau \rbrace$. Next, let $\hat{S}_{\tau}=\hat{S}_{\tau}(T)\subseteq \mathcal{A}$ denote the recommendation of a learning algorithm (under consideration) after $T$ time units of exploration, while $\hat{S}_{\tau}^c$ denotes its complement.
The performance of the learning agent is measured by the accuracy with which it can classify the arms into $S_{\tau}$ and $S_{\tau}^{c}$ after time horizon $T$. Equivalently, using $\mathbb{I}(E)$ to denote the indicator of an event $E$, the \emph{loss} $\mathcal{L}(T)$ is defined as
\begin{align*}
\Ls (T) = \mathbb{I}\big(\lbrace S_{\tau}\cap \hat{S}_{\tau}^{c}\neq \emptyset\rbrace    \cup    \lbrace\hat{S}_{\tau}\cap S_{\tau}^{c}\neq \emptyset\rbrace\big).
\end{align*}			
Finally, the goal of the learning agent is to minimize the expected loss:
\begin{align*}
\E[\Ls(T)] = \Pb\big(\lbrace S_{\tau}\cap \hat{S}_{\tau}^{c} \neq \emptyset \rbrace  \cup   \lbrace \hat{S}_{\tau}\cap S_{\tau}^{c} \neq \emptyset\rbrace\big).
\end{align*}
Note that the expected loss is simply the \emph{probability of mis-classification} (i.e., error), that occurs either if a good arm is rejected or a bad arm is accepted as a good one.




The above TBP formulation has several applications, for instance, from areas ranging from anomaly detection and classification (see  \citet{locatelli2016optimal}) to industrial applications as well as in mobile communications (see \citet{audibert2010best})  where the users are to be allocated only those channels whose quality is above an acceptable threshold.

\subsection{Related Work}
\label{prevRes}
Significant amount of literature is available on the stochastic MAB setting with respect to minimizing the cumulative regret. While the seminal work of \cite{robbins1952some}, \cite{thompson1933likelihood},  and \cite{lai1985asymptotically} prove asymptotic lower bounds on the cumulative regret, the more recent work of \cite{auer2002finite} propose the UCB1 algorithm that provides finite time-horizon guarantees. 
%
%
Subsequent work such as \cite{audibert2009minimax} and \cite{auer2010ucb} have improved the upper bounds on the cumulative regret. The authors in \cite{auer2010ucb} have proposed a \emph{round-based}\footnote{An algorithm is said to be \textit{round-based} if it pulls all the arms equal number of times in each round, and then proceeds to eliminate one or more arms that it identifies to be sub-optimal.} version of the UCB algorithm, referred to as UCB-Improved. Of special mention is the work of \cite{audibert2009exploration} where the authors have introduced a \emph{variance-aware} UCB algorithm, referred to as UCB-V; it is shown that the algorithms that take into account variance estimation along with mean estimation tends to perform better than the algorithms that solely focuses on mean estimation, for instance, such as UCB1.
For a more detail survey of literature on UCB algorithms, we refer the reader to \cite{bubeck2012regret}.

	

In this work we are particularly interested in \emph{pure-exploration MABs},  where the focus in primarily on simple regret rather than the cumulative regret. The relationship between cumulative regret and simple regret is proved in \cite{bubeck2011pure} where the authors prove that minimizing the simple regret necessarily results in maximizing the cumulative regret.
The pure exploration problem has been explored  mainly under the following two settings:
	
	\emph{1. Fixed Budget setting:} Here the learning algorithm has to suggest the best arm(s) within a fixed time-horizon $T$, that is usually given as an input. The objective is to maximize the probability of returning the best arm(s).  This is the scenario we consider in our paper. In \cite{audibert2010best} the authors propose the  UCBE and the Successive Reject (SR) algorithm, and prove simple-regret guarantees for the problem of identifying the single best arm.  In the combinatorial fixed budget setup \cite{gabillon2011multi} propose the GapE and GapE-V algorithms that suggest, with high probability, the best $m$ 
	 arms at the end of the time budget. Similarly, \cite{bubeck2013multiple} introduce the  Successive Accept Reject (SAR) algorithm, which is an extension of the SR algorithm; SAR is a round based algorithm whereby at the end of each round an arm is either accepted or rejected (based on certain confidence conditions) until the top $m$ arms are suggested at the end of the budget with high probability. A similar combinatorial setup was explored in \cite{chen2014combinatorial} where the authors propose the Combinatorial Successive Accept Reject (CSAR) algorithm, which is similar in concept to SAR but with a more general setup. 

	\emph{2. Fixed Confidence setting:} In this setting the learning algorithm has to suggest the best arm(s) with a fixed confidence (given as input) with as fewer number of attempts as possible. The single best arm identification has been studied in \cite{even2006action}, while for the combinatorial setup \cite{kalyanakrishnan2012pac} have proposed the LUCB algorithm which, on termination, returns  $m$ arms which are at least $\epsilon$ close to the true top-$m$ arms with probability at least $1-\delta$. For a detail survey of this setup we refer the reader to \cite{jamieson2014best}. 

Apart from these two settings some unified approaches has also been suggested in \cite{gabillon2012best} which proposes the algorithms UGapEb and UGapEc which can work in both the above two settings. The thresholding bandit problem is a specific instance of the pure-exploration setup of \cite{chen2014combinatorial}. In the latest work of \cite{locatelli2016optimal} Anytime Parameter-Free Thresholding (APT) algorithm comes up with an improved anytime guarantee than CSAR for the thresholding bandit problem.

\subsection{Our Contribution}
\label{contribution}
In this paper we propose the Augmented UCB (AugUCB) algorithm for the fixed-budget setting of a specific combinatorial, pure-exploration, stochastic MAB called the thresholding bandit problem.
 AugUCB essentially combines the approach of UCB-Improved, CCB \citep{liu2016modification} and APT algorithms. Our algorithm takes into account the empirical variances of the arms along with mean estimates; to the best of our knowledge this is the first variance-based algorithm for the considered TBP. 
Thus, we also address an open problem discussed in \cite{auer2010ucb} of designing an algorithm that can eliminate arms based on variance estimates. In this regard, note that both CSAR and APT are not variance-based algorithms. 

\begin{table}[b]
\caption{AugUCB vs.\ State of the art}
\label{tab:regret-bds}
\begin{center}
\begin{tabular}{|p{1.3cm}|p{6.4cm}|}
\hline
Algorithm  & Upper Bound on Expected Loss \\
\hline
\hline
AugUCB      &$ \exp\left(- \frac{T}{4096 \log(K\log K)H_{\sigma,2}} + \log\left(2KT\right) \right) $ \\
\hline
\hline
UCBEV		&$\exp\left(-\frac{1}{512}\frac{T-2K}{H_{\sigma,1}} + \log\left(6KT\right)\right)$ \\
\hline
\hline
APT         &$\exp\left(-\frac{T}{64 H_1}+2\log((\log(T)+1)K)\right)$ \\
\hline
\hline
CSAR		&$\exp\left(-\frac{T-K}{72\log(K)H_{CSAR,2}}+2\log(K)\right)$ \\
\hline

\end{tabular}
\end{center}
\end{table}

Our theoretical contribution comprises 
 proving an upper bound on the expected loss incurred by AugUCB (Theorem~\ref{Result:Theorem:1}).
In Table \ref{tab:regret-bds} we compare the upper bound on the losses incurred by the various algorithms, including AugUCB. The terms $H_1, H_2$, $H_{CSAR,2}, H_{\sigma,1}$ and $H_{\sigma,2}$ represent various problem complexities, and are as defined in Section~\ref{results}. From Section~\ref{results} we note that, for all $K\ge8$, we have
\begin{align*}
\log\left(K\log K\right) H_{\sigma,2} > \log(2K) H_{\sigma,2} \ge H_{\sigma,1}.
\end{align*}
Thus, it follows that the upper bound for UCBEV is better than that for AugUCB.
 However, implementation of UCBEV algorithm requires $H_{\sigma,1}$ as input, whose computation is not realistic in practice. In contrast, our AugUCB algorithm requires no such complexity factor as input. 

Proceeding with the comparisons, we emphasize that the upper bound for  AugUCB is, in fact, not comparable with that of APT and CSAR; this is because the complexity term $H_{\sigma,2}$ is not explicitly comparable with either $H_1$ or $H_{CSAR,2}$. However, through extensive simulation experiments we find that AugUCB significantly outperforms both APT, CSAR and other non variance-based algorithms. AugUCB also outperforms UCBEV under explorations where non-optimal values of $H_{\sigma,1}$  are used. In particular, we consider experimental scenarios comprising large number of arms, with the variances of arms in $S_\tau$ being large. AugUCB, being variance based, exhibits superior performance under these settings.  
%

%

The remainder of the paper is organized as follows. In section \ref{algorithm} we present our AugUCB algorithm. 
Section \ref{results} contains our main theorem on expected loss, while section \ref{expt} contains simulation experiments. We finally draw our conclusions in section \ref{conclusion}.
%
%
%
\vspace*{-1em}
\section{Augmented-UCB Algorithm}
\label{algorithm}

\algblock{ArmElim}{EndArmElim}
\algnewcommand\algorithmicArmElim{\textbf{\em Arm Elimination by Mean Estimation}}
 \algnewcommand\algorithmicendArmElim{}
\algrenewtext{ArmElim}[1]{\algorithmicArmElim\ #1}
\algrenewtext{EndArmElim}{\algorithmicendArmElim}
\algtext*{EndArmElim}

\algblock{ArmElimV}{EndArmElimV}
\algnewcommand\algorithmicArmElimV{\textbf{\em Arm Elimination by Mean and Variance Estimation}}
 \algnewcommand\algorithmicendArmElimV{}
\algrenewtext{ArmElimV}[1]{\algorithmicArmElimV\ #1}
\algrenewtext{EndArmElimV}{\algorithmicendArmElimV}
\algtext*{EndArmElimV}

\algblock{ResetParam}{EndResetParam}
\algnewcommand\algorithmicResetParam{\textbf{\em Reset Parameters}}
 \algnewcommand\algorithmicendResetParam{}
\algrenewtext{ResetParam}[1]{\algorithmicResetParam\ #1}
\algrenewtext{EndResetParam}{\algorithmicendResetParam}
\algtext*{EndResetParam}

\label{notation}
\textbf{Notation and assumptions:} $\mathcal{A}$ denotes the set of arms, and $|\mathcal{A}|=K$ is the number of arms in $\mathcal{A}$. 
For arm $i\in\mathcal{A}$, we use $r_{i}$ to denote the true mean of the distribution from which the rewards are sampled, while $\hat{r}_{i}(t)$ denotes the estimated mean at time $t$. Formally, using $n_i(t)$ to denote the number of times arm $i$ has been pulled until time $t$, we have $\hat{r}_{i}(t)=\frac{1}{n_{i}(t)}\sum_{z=1}^{n_i(t)} X_{i,z}$, where $X_{i,z}$ is the reward sample received when arm $i$ is pulled for the $z$-th time. %
Similarly, we use $\sigma_{i}^{2}$ to denote the true variance of the reward distribution corresponding to arm $i$, while $\hat{v}_{i}(t)$ is the estimated variance, i.e., $\hat{v}_{i}(t)=\frac{1}{n_i(t)}\sum_{z=1}^{n_{i}(t)}(X_{i,z}-\hat{r}_{i})^{2}$. Whenever there is no ambiguity about the underlaying  time index $t$, for simplicity we neglect $t$ from the notations and simply use  $\hat{r}_i, \hat{v}_i,$ and $n_i, $ to denote the respective quantities.  Let  $\Delta_{i}=|\tau-r_{i}|$ denote the distance of the true mean from the threshold $\tau$. Also, the rewards are assumed to take values in $[0,1]$.


\textbf{The Algorithm:} The Augmented-UCB (AugUCB) algorithm is presented in Algorithm~\ref{alg:augucb}.
AugUCB is essentially based on the arm elimination method of the UCB-Improved \cite{auer2010ucb}, but adapted to the thresholding bandit setting proposed in \cite{locatelli2016optimal}. However, unlike the UCB improved (which is based on mean estimation) our algorithm employs \emph{variance estimates} (as in \cite{audibert2009exploration}) for arm elimination; to the best of our knowledge this is the first variance-aware  algorithm for the thresholding bandit problem. Further, we allow for arm-elimination at each time-step, which is in contrast to the earlier work (e.g., \cite{auer2010ucb,chen2014combinatorial}) where the arm elimination task is deferred to the end of the respective exploration rounds. The details are presented below.

The active set $B_{0}$ is initialized with all the arms from $\mathcal{A}$. We divide the entire budget $T$ into rounds/phases like in UCB-Improved, CCB, SAR and CSAR. At every time-step AugUCB checks for arm elimination conditions, while updating parameters at the end of each round. As suggested by \cite{liu2016modification} to make AugUCB to overcome too much early exploration, we no longer pull all the arms equal number of times in each round. Instead, we choose an arm in the active set $B_m$ that minimizes $(|\hat{r}_{i} - \tau |-2s_i)$ where 
\begin{small}
\begin{align*}
s_i & = \sqrt{\frac{\rho\psi_m (\hat{v}_{i}+1) \log ( T \epsilon_{m})}{4 n_{i}}} 
\end{align*}
\end{small} 
with $\rho$ being the arm elimination parameter and $\psi_{m}$ being the exploration regulatory factor.
The above condition ensures that an arm closer to the threshold $\tau$ is pulled; 
parameter $\rho$ can be used to fine tune the elimination interval.
The choice of exploration factor, $\psi_m$,
comes directly from \cite{audibert2010best} and \cite{bubeck2011pure} where it is  stated that in pure exploration setup, the exploring factor must be linear in $T$ (so that an exponentially small probability of error is achieved) rather than being logarithmic in $T$ (which is more suited for minimizing cumulative regret).

\begin{algorithm}[t!]
\caption{AugUCB}
\label{alg:augucb}
\begin{algorithmic}
\State {\bf Input:} Time budget $T$; parameter $\rho$; 
  threshold $\tau$
\State {\bf Initialization:} $B_{0}=\mathcal{A}$; $m=0$; $\epsilon_{0}=1$;
\begin{small}
\begin{align*}
M&=\left\lfloor \frac{1}{2}\log_{2} \frac{T}{e}\right\rfloor; 
\hspace{2mm}\psi_{0}=\frac{T\epsilon_{0}}{128\Big(\log(\frac{3}{16}K\log K)\Big)^2}; \\
\ell_{0}&=\left\lceil \frac{2\psi_0\log( T\epsilon_{0})}{\epsilon_{0}} \right\rceil;
\hspace{2mm}N_{0}=K\ell_{0}
\end{align*}
\end{small}
\State Pull each arm once
\vspace{-2mm}
\State \For{$t=K+1,..,T$}
\State Pull arm $j\in\argmin_{i\in B_{m}}\Big\lbrace |\hat{r}_{i} - \tau | - 2s_{i}\Big\rbrace$
\vspace{-4mm}
\State \For{$i\in B_m$}
\vspace{-4mm}
\State \If{$(\hat{r}_{i} + s_i  < \tau - s_i)$ or $(\hat{r}_{i} - s_i > \tau + s_i)$}
\State $B_m\leftarrow B_m\backslash\{i\}$\hspace{4mm} (Arm deletion)
\EndIf
\EndFor
\vspace{-2mm}
\State \If{$t\geq N_{m}$ and $m \leq M$}
\State \textbf{Reset Parameters}
\State $\epsilon_{m+1}\leftarrow\frac{\epsilon_{m}}{2}$
\State $B_{m+1} \leftarrow B_{m}$
\State $\psi_{m+1}\leftarrow \frac{T\epsilon_{m+1}}{128(\log(\frac{3}{16}K\log K))^{2}}$
\State $\ell_{m+1}\leftarrow\left\lceil \frac{2\psi_{m+1}\log( T\epsilon_{m+1})}{\epsilon_{m+1}} \right\rceil$
\State $N_{m+1} \leftarrow t + |B_{m+1}|\ell_{m+1}$
\State $m \leftarrow m+1$
\EndIf
\EndFor
\State \textbf{Output:} $\hat{S}_{\tau}=\lbrace i: \hat{r}_{i}\geq \tau \rbrace$.
\end{algorithmic}
\end{algorithm}


%
\vspace{-2mm}
\section{Theoretical Results}
\label{results}

Let us begin by recalling the following definitions of the  \emph{problem complexity} as introduced in \cite{locatelli2016optimal}:
\begin{align*}
H_{1} = \sum_{i=1}^{K}\dfrac{1}{\Delta_{i}^{2}} \hspace{1mm}\text{     and }  \hspace{1mm}
H_{2} =\min_{i\in \mathcal{A}}\dfrac{i}{{\Delta_{(i)}^{2}}} 
\end{align*}
where $(\Delta_{(i)}: i\in\mathcal{A})$ is obtained by arranging $(\Delta_i:i\in\mathcal{A})$ in an increasing order. Also, from \cite{chen2014combinatorial} we have
\begin{align*}
H_{CSAR,2}=\max_{i\in\mathcal{A}}\frac{i}{\Delta_{(i)}^2}.
\end{align*}
$H_{CSAR,2}$ is the complexity term appearing in the bound for the CSAR algorithm. The relation between the above complexity terms are as follows (see \cite{locatelli2016optimal}):
%
\begin{align*}
H_{1}\leq \log(2K)H_{2} \mbox{ and }
 H_1 \leq \log(K)H_{CSAR,2}.
\end{align*}

As ours is a variance-aware algorithm, we require $H_{1}^{\sigma}$ (as defined in \cite{gabillon2011multi}) that incorporates reward variances into its expression as given below:
\begin{align*}
 H_{\sigma,1}=\sum_{i=1}^{K}\frac{\sigma_{i}+\sqrt{\sigma_{i}^{2}+(16/3)\Delta_{i}}}{\Delta_{i}^{2}}.
\end{align*}
Finally, analogous to $H_{CSAR,2}$, in this paper we introduce the complexity term $H_{\sigma,2}$, which is given by
\begin{align*}
H_{\sigma,2}=\max_{i\in \mathcal{A}} \frac{i}{\tilde{\Delta}_{(i)}^{2}}
\end{align*}
where $\tilde{\Delta}_{i}^{2}=\frac{\Delta_{i}^{2}}{\sigma_{i}+\sqrt{\sigma_{i}^{2}+(16/3)\Delta_{i}}}$, and $(\tilde{\Delta}_{(i)})$ is an increasing ordering of $(\tilde{\Delta}_{i})$. Following the results in \cite{audibert2010best}, we can show that
\begin{align*}
H_{\sigma,2}\le H_{\sigma,1}\le\overline{\log}(K) H_{\sigma,2} \le \log(2K) H_{\sigma,2}.
\end{align*}

%

Our main result is summarized in the following theorem where we prove an  upper bound on the expected loss. 
\begin{theorem}
\label{Result:Theorem:1}
For $K\geq 4$ and
$\rho={1}/{3}$,
the expected loss of the AugUCB algorithm is given by,
\begin{align*}
\E[\Ls(T)]
& \leq 2KT
 \exp\bigg(- \frac{T}{4096 \log( K\log K) H_{\sigma,2}} \bigg).
\end{align*}
\end{theorem}

\begin{proof}
The proof comprises of two modules. In the first module we investigate the necessary conditions for arm elimination within a specified number of rounds, which is motivated by the technique in \cite{auer2010ucb}. Bounds on the arm-elimination probability is then obtained; however, since we use variance estimates, we invoke the Bernstein inequality (as in \cite{audibert2009exploration}) rather that the Chernoff-Hoeffding bounds (which is appropriate for the UCB-Improved \citep{auer2010ucb}). In the second module, as in \cite{locatelli2016optimal}, we first define a favourable event that will yield an upper bound on the expected loss. Using union bound, we then incorporate the result from module-1 (on the arm elimination probability), and finally derive the result through a series of simplifications.
The details are as follows.

\textbf{Arm Elimination:} Recall the notations used in the algorithm, Also, for each arm $i\in\mathcal{A}$, define $m_{i}=\min\left\lbrace m| \sqrt{\rho\epsilon_{m}}<\frac{\Delta_{i}}{2}\right\rbrace$. In the $m_i$-th round, whenever $n_i=\ell_{m_i}\ge\frac{2\psi_{m_i}\log{(T\epsilon_{m_{i}})}}{\epsilon_{m_{i}}}$, we obtain (as $\hat{v}_i\in[0,1]$)
%
%
\begin{align}
\label{si_bound_equn}
s_i 
&\le \sqrt{\frac{\rho(\hat{v}_i+1)\epsilon_{m_i}}{8}}
  \le \frac{\sqrt{\rho\epsilon_{m_i}}}{2} < \frac{\Delta_i}{4}.
\end{align}

First, let us consider a bad arm $i\in\mathcal{A}$ (i.e., $r_i<\tau$). We note that, in the $m_i$-th round  whenever 
$\hat{r}_i \le r_i +2s_i$, then arm $i$ is eliminated as a bad arm. This is easy to verify as follows: using (\ref{si_bound_equn}) we obtain,
\begin{align*}
\hat{r}_{i}\leq r_{i} + 2s_{i} 
< r_{i} + \Delta_{i} - 2s_{i} 
= \tau - 2s_{i} 
\end{align*}
which is precisely one of the elimination conditions in Algorithm~\ref{alg:augucb}. Thus, the probability that a bad arm is not eliminated correctly in the $m_i$-th round (or before) is given by

\noindent
\begin{align}
\mathbb{P}(\hat{r}_{i}> r_{i} + 2s_{i})
&\leq \mathbb{P}\left( \hat{r}_{i} > r_{i}+ 2\bar{s}_i\right)  
+ \mathbb{P}\left( \hat{v}_{i}\geq \sigma_{i}^{2}+\sqrt{\rho\epsilon_{m_{i}}}\right)\label{eq:prob_eq2}
\end{align}
where 
\begin{align*}
\bar{s}_i=\sqrt{\dfrac{\rho\psi_{m_i} (\sigma_{i}^{2}+\sqrt{\rho\epsilon_{m_{i}}} + 1)\log( T\epsilon_{m_{i}})}{4n_{i}}}
\end{align*}
Note that, substituting $n_i=\ell_{m_i}\ge \frac{2\psi_{m_i}\log{(T\epsilon_{m_{i}})}}{\epsilon_{m_{i}}}$, $\bar{s}_i$ can be simplified to obtain,
\begin{align}
2\bar{s}_i
\leq \dfrac{\sqrt{\rho\epsilon_{m_{i}}(\sigma_{i}^{2}+\sqrt{\rho\epsilon_{m_{i}}} + 1)}}{2}\leq \sqrt{\rho \epsilon_{m_{i}}}.
\label{si_bar_equn}
\end{align}


The first term in the LHS of (\ref{eq:prob_eq2}) can be bounded using the Bernstein inequality as below:
\begin{align}
&\mathbb{P}\left( \hat{r}_{i} > r_{i}+ 2\bar{s}_i\right)\nonumber \\
&\le \exp\left(- \dfrac{(2\bar{s}_i)^2 n_i}{2\sigma_i^2+\frac{4}{3}\bar{s}_i}\right)\nonumber \\
& \le \exp\left(- \dfrac{\rho\psi_{m_i} (\sigma_{i}^{2}+\sqrt{\rho\epsilon_{m_{i}}} + 1)\log( T\epsilon_{m_{i}})}{2\sigma_i^2+\frac{2}{3}\sqrt{\rho \epsilon_{m_{i}}}}\right)\nonumber \\
& \overset{(a)}{\leq} \exp\left(- \dfrac{3\rho T\epsilon_{m_i}}{256 a^2} \left(\dfrac{\sigma_{i}^{2}+\sqrt{\rho\epsilon_{m_{i}}}+1}{3\sigma_{i}^{2}+\sqrt{\rho \epsilon_{m_{i}}}}\right) \log( T\epsilon_{m_{i}}) \right) \nonumber \\
&:= \exp(-Z_i) 
\label{lhs1_equn}
\end{align}
where, for simplicity, we have used $\alpha_i$ to denoted the exponent in the inequality $(a)$.
Also, note that $(a)$ is obtained by using  $\psi_{m_i}=\frac{T\epsilon_{m_i}}{128a^{2}}$, where $a=(\log(\frac{3}{16} K\log K))$.

 The second term in the LHS of (\ref{eq:prob_eq2}) can be simplified as follows:
\begin{align}
&\mathbb{P}\bigg\lbrace \hat{v}_{i}\geq \sigma_{i}^{2}+\sqrt{\rho\epsilon_{m_{i}}}\bigg\rbrace\nonumber\\
&\leq \mathbb{P}\bigg\lbrace \dfrac{1}{n_{i}}\sum_{t=1}^{n_{i}}(X_{i,t}-r_{i})^{2}-(\hat{r}_{i}-r_{i})^{2}\geq \sigma_{i}^{2}+\sqrt{\rho\epsilon_{m_{i}}}\bigg\rbrace\nonumber\\
&\leq \mathbb{P}\bigg\lbrace \dfrac{\sum_{t=1}^{n_{i}}(X_{i,t}-r_{i})^{2}}{n_{i}}\geq \sigma_{i}^{2}+\sqrt{\rho\epsilon_{m_{i}}} \bigg\rbrace\nonumber\\
&\overset{(a)}{\leq} \mathbb{P}\bigg\lbrace \dfrac{\sum_{t=1}^{n_{i}}(X_{i,t}-r_{i})^{2}}{n_{i}}\geq \sigma_{i}^{2} + 2\bar{s}_i\bigg\rbrace \nonumber\\
&\overset{(b)}{\leq} \exp\bigg(- \dfrac{3\rho\psi_{m_i}}{2} \bigg(\dfrac{\sigma_{i}^{2}+\sqrt{\rho\epsilon_{m_{i}}}+1}{3\sigma_{i}^{2}+\sqrt{\rho \epsilon_{m_{i}}}}\bigg) \log( T\epsilon_{m_{i}}) \bigg)\nonumber \\
& = \exp(-Z_i)
\label{lhs2_equn}
\end{align}
where inequality $(a)$ is obtained using (\ref{si_bar_equn}), while $(b)$ follows from the Bernstein inequality. 
  
Thus, using (\ref{lhs1_equn}) and (\ref{lhs2_equn}) in (\ref{eq:prob_eq2}) we obtain $\mathbb{P}(\hat{r}_{i}> r_{i} + 2s_{i})\le 2\exp(-Z_i)$.
 %
Proceeding similarly, for a good arm $i\in\mathcal{A}$, the probability that it is not correctly eliminated in the $m_i$-th round (or before) is also bounded by $\mathbb{P}(\hat{r}_{i}< r_{i} - 2s_{i})\le 2\exp(-Z_i)$. In general, for any $i\in\mathcal{A}$ we have
\begin{align}
\Pb(|\hat{r}_i-r_i|>2s_i) 
&\le4\exp(-Z_i).
\label{final_bound_equn}
\end{align}


\textbf{Favourable Event:} Following the notation in \cite{locatelli2016optimal} we define the event
\begin{align*}
\xi&=\bigg\lbrace \forall i\in \mathcal{A}, \forall t=1,2,..,T: |\hat{r_i} - r_i| \leq  2s_i\bigg\rbrace.
\end{align*}
Note that, on $\xi$ each arm $i\in \mathcal{A}$  is eliminated correctly in the $m_i$-th round (or before). Thus, it follows that $\mathbb{E}[\mathcal{L}(T)]\le P(\xi^c)$. Since $\xi^c$ can be expressed as an union of the events $(|\hat{r}_i-r_i|>2s_i)$ for all $i\in\mathcal{A}$ and all $t=1,2,\cdots,T$, using union bound we can write
\begin{align*}
&\mathbb{E}[\mathcal{L}(T)] 
\le \sum_{i\in\mathcal{A}}\sum_{t=1}^T \Pb(|\hat{r}_i-r_i|>2s_i) \\
&\le \sum_{i\in\mathcal{A}}\sum_{t=1}^T 4 \exp(-Z_i) \\
&\le 4T\sum_{i\in\mathcal{A}} \exp\left(- \dfrac{3\rho T\epsilon_{m_i}}{256 a^2} \left(\dfrac{\sigma_{i}^{2}+\sqrt{\rho\epsilon_{m_{i}}}+1}{3\sigma_{i}^{2}+\sqrt{\rho \epsilon_{m_{i}}}}\right) \log( T\epsilon_{m_{i}}) \right) \\
&\overset{(a)}{\le} 4T \sum_{i\in\mathcal{A}} \exp\left(- \frac{3T\Delta_{i}^{2}}{4096 a^2} \left(\frac{4\sigma_{i}^{2}+\Delta_{i}+4}{12\sigma_{i}^{2}+\Delta_{i}}\right) \log( \frac{3}{16} T\Delta_{i}^{2}) \right) \\
&\overset{(b)}{\le} 4T \sum_{i\in\mathcal{A}}\exp\bigg(- \frac{12T\Delta_{i}^{2}}{(12\sigma_{i}+ 12\Delta_{i})}\frac{\log (\frac{3}{16} K\log K)}{4096 a^2 } \bigg) \\
&\overset{(c)}{\le} 4T \sum_{i\in\mathcal{A}} \exp\bigg(- \frac{T\Delta_{i}^{2}\log ( \frac{3}{16} K\log K)}{4096 (\sigma_{i} + \sqrt{\sigma_{i}^{2} + (16/3)\Delta_{i}}) a^2} \bigg) \\
& \overset{(d)}{\le} 4T \sum_{i\in\mathcal{A}} \exp\bigg(- \frac{T\log ( \frac{3}{16} K\log K)}{4096 \tilde{\Delta}_i^{-2} a^2} \bigg) \\
& \overset{(e)}{\le}4T \sum_{i\in\mathcal{A}} \exp\bigg(- \frac{T\log ( \frac{3}{16} K\log K)}{4096 \max_{j}(j\tilde{\Delta}_{(j)}^{-2}) (\log(\frac{3}{16} K\log K))^{2}} \bigg) \\
& \overset{(f)}{\le}4KT \exp\bigg(- \frac{T}{4096 \log(K\log K)H_{\sigma,2}}\bigg).
\end{align*}
The justification for the above simplifications are as follows:

\noindent
 $\bullet$ $(a)$ is obtained by noting that in round $m_i$ we have 
 $\frac{\Delta_i}{4}\leq\sqrt{\rho\epsilon_{m_{i}}}<\frac{\Delta_i}{2}.$

\noindent
 $\bullet$ For $(b)$, we note that the function $x\mapsto x\exp(-Cx^2)$, where $x\in[0,1]$, is  decreasing on $[1/\sqrt{2C},1]$ for any $C>0$ (see \cite{bubeck2011pure,auer2010ucb}). Thus, using $C=\lfloor T/e\rfloor$ and $\min_{j\in \mathcal{A}}\Delta_j =\Delta =\sqrt{\frac{K\log K}{T}} > \sqrt{\frac{e}{T}}$,
we obtain (b).

\noindent
 $\bullet$
To obtain $(c)$ we have used the inequality $\Delta_i\le \sqrt{\sigma_{i}^{2} + (16/3)\Delta_{i}}$ (which holds because $\Delta_i\in[0,1]$).

\noindent
 $\bullet$
 $(d)$ is obtained simply by substituting $\tilde{\Delta}_i=\frac{\Delta_{i}^{2}}{\sigma_{i}+\sqrt{\sigma_{i}^{2}+(16/3)\Delta_{i}}}$ and $a=\log(\frac{3}{16} K\log K)$.
 
 \noindent
 $\bullet$
 Finally, to obtain $(e)$ and $(f)$, note that 
$\tilde{\Delta}_i^{-2}\le i\tilde{\Delta}_i^{-2} \le \max_{j\in\mathcal{A}}j\Delta_{(j)}^{-2}=H_{\sigma,2}.$
\end{proof}

\vspace{-4mm}
\section{Numerical Experiments}
\label{expt}

In this section, we empirically compare the  performance of AugUCB against APT, UCBE, UCBEV, CSAR and the uniform-allocation (UA) algorithms. A brief note about these algorithms are as follows:

$\bullet$ APT: This algorithm is from \cite{locatelli2016optimal}; we set $\epsilon=0.05$, which is the margin-of-error within which APT suggests the set of good arms.

$\bullet$ AugUCB: This is the Augmented-UCB algorithm proposed in this paper; as in Theorem \ref{Result:Theorem:1} we set $\rho=\frac{1}{3}$.

$\bullet$ UCBE: This is a modification of the algorithm in \cite{audibert2009exploration} (as it was originally proposed for the best arm identification problem); here, we set $a=\frac{T-K}{H_1}$, and at each time-step an arm $i\in\argmin\left\lbrace |\hat{r}_{i} -\tau|-\sqrt{\frac{a}{n_{i}}} \right\rbrace$ is pulled.

$\bullet$ UCBEV: This is a modification of the algorithm in \cite{gabillon2011multi} (proposed for the TopM problem); its implementation is identical to UCBE, but with $a = \frac{T-2K}{H_{\sigma,1}}$. As mentioned earlier, note that UCBEV's implementation would not be possible in real scenarios, as it requires computing the problem complexity $H_{\sigma,1}$. However, for theoretical reasons we show the best performance achievable by UCBEV. In experiment 6 we perform further explorations of UCBEV with alternate settings of $a$.

$\bullet$ CSAR:  Modification of the successive-reject algorithm in \cite{chen2014combinatorial}; here, we reject the arm farthest from $\tau$ after each round. 

$\bullet$ UA: The naive strategy where at each time-step an arm is uniformly sampled from $\mathcal{A}$; however, UA is known to be optimal if all arms are equally difficult to classify. 

\noindent
Motivated by the settings considered in \cite{locatelli2016optimal}, 
we design six different experimental scenarios that are obtained by varying the arm means and variances. 
Across all experiments consists of $K=100$  arms (indexed $i=1,2,\cdots,100$) of which ${S}_\tau=\{6,7,\cdots,10\}$, where we have fixed $\tau=0.5$.
In all the experiments, each algorithm is run independently for $10000$ time-steps. At every time-step, the output set,  $\hat{S}_\tau$, suggested by each algorithm is recorded; the output is counted as an error if $\hat{S}_\tau\ne S_\tau$. In Figure~1, for each experiment, we have reported the percentage of error incurred by the different algorithms as a function of time; Error percentage is obtained by repeating each experiment independently  for $500$ iterations, and then respectively computing the fraction of errors. The details of the considered experiments are as follows.


 


	
\textbf{Experiment-1:} The reward distributions are Gaussian with  means  $r_{1:4}=0.2+(0:3)\cdot0.05$, $r_{5}=0.45$, $r_{6}=0.55$, $r_{7:10}=0.65+(0:3)\cdot0.05$ and $r_{11:100}=0.4$. Thus, the means of the first $10$ arms follow an arithmetic progression. The remaining arms have identical means; this setting is chosen because now a significant budget is required in exploring these arms, thus increasing the problem complexity.

\begin{figure}[t]
    \centering
    \begin{tabular}{cc}
    \subfigure[0.32\textwidth][Expt-$1$: Arithmetic Progression (Gaussian)]
    {
    		\pgfplotsset{
		tick label style={font=\Huge},
		label style={font=\Huge},
		legend style={font=\Large},
		}
        \begin{tikzpicture}[scale=0.4]
      	\begin{axis}[
		xlabel={Time-step},
		ylabel={Error Percentage},
		grid=major,
        clip=true,
  		legend style={at={(0.5,1.2)},anchor=north, legend columns=3} ]
		\addplot table{results/budgetTestAP/APT12_comp_subsampled.txt};
		\addplot table{results/budgetTestAP/AugUCBV1_comp_subsampled.txt};
		\addplot table{results/budgetTestAP/UCBEM1_comp_subsampled.txt};
		\addplot table{results/budgetTestAP/UCBEMV1_comp_subsampled.txt};
		\addplot table{results/budgetTestAP/SR1_comp_subsampled.txt};
		\addplot table{results/budgetTestAP/UA1_comp_subsampled.txt};
      	\legend{APT,AugUCB,UCBE,UCBEV,CSAR,UA}
      	\end{axis}
      	\end{tikzpicture}
  		\label{Fig:budgetExpt1}
    }
    &
    \subfigure[0.32\textwidth][Expt-$2$: Geometric Progression (Gaussian)]
    {
    	\pgfplotsset{
		tick label style={font=\Huge},
		label style={font=\Huge},
		legend style={font=\Large},
		}
        \begin{tikzpicture}[scale=0.4]
        \begin{axis}[
		xlabel={Time-step},
		ylabel={Error Percentage},
		grid=major,
		clip=true,
  		legend style={at={(0.5,1.2)},anchor=north, legend columns=3} ]
		\addplot table{results/budgetTestGP/APT12_comp_subsampled.txt};
		\addplot table{results/budgetTestGP/AugUCBV1_comp_subsampled.txt};
		\addplot table{results/budgetTestGP/UCBEM1_comp_subsampled.txt};
		\addplot table{results/budgetTestGP/UCBEMV1_comp_subsampled.txt};
		\addplot table{results/budgetTestGP/SR1_comp_subsampled.txt};
		\addplot table{results/budgetTestGP/UA1_comp_subsampled.txt};
        \legend{APT,AugUCB,UCBE,UCBEV,CSAR,UA}
      	\end{axis}
      	\label{Fig:budgetExpt2}
        \end{tikzpicture}
    }
    \\
    \subfigure[0.32\textwidth][Expt-$3$: Three Group Setting (Gaussian)]
    {
    		\pgfplotsset{
		tick label style={font=\Huge},
		label style={font=\Huge},
		legend style={font=\Large},
		}
        \begin{tikzpicture}[scale=0.4]
        \begin{axis}[
		xlabel={Time-step},
		ylabel={Error Percentage},
       	grid=major,
       	clip=true,
  		legend style={at={(0.5,1.2)},anchor=north, legend columns=3} ]
		\addplot table{results/budgetTestGR1/APT1_comp_subsampled.txt};
		\addplot table{results/budgetTestGR1/AugUCB1_comp_subsampled.txt};
		\addplot table{results/budgetTestGR1/UCBEM1_comp_subsampled.txt};
		\addplot table{results/budgetTestGR1/UCBEMV1_comp_subsampled.txt};
		\addplot table{results/budgetTestGR1/SR1_comp_subsampled.txt};
		\addplot table{results/budgetTestGR1/UA1_comp_subsampled.txt};
        \legend{APT,AugUCB,UCBE,UCBEV,CSAR,UA}
      	\end{axis}
      	\end{tikzpicture}
   		\label{Fig:budgetExpt3} 
    }
    &
    \subfigure[0.32\textwidth][Expt-$4$: Two Group Setting (Gaussian) ]
    {
    	\pgfplotsset{
		tick label style={font=\Huge},
		label style={font=\Huge},
		legend style={font=\Large},
		}
        \begin{tikzpicture}[scale=0.4]
        \begin{axis}[
		xlabel={Time-step},
		ylabel={Error Percentage},
		grid=major,
		clip=true,
  		legend style={at={(0.5,1.2)},anchor=north, legend columns=3} ]
		\addplot table{results/budgetTestGR2/APT1_comp_subsampled.txt};
		\addplot table{results/budgetTestGR2/AugUCBV1_comp_subsampled.txt};
		\addplot table{results/budgetTestGR2/UCBEM1_comp_subsampled.txt};
		\addplot table{results/budgetTestGR2/UCBEMV1_comp_subsampled.txt};
		\addplot table{results/budgetTestGR2/SR1_comp_subsampled.txt};
		\addplot table{results/budgetTestGR2/UA1_comp_subsampled.txt};
        \legend{APT,AUgUCB,UCBE,UCBEV,CSAR,UA}
      	\end{axis}
      	\label{Fig:budgetExpt4}
        \end{tikzpicture}
    }
    \\
    \subfigure[0.32\textwidth][Expt-$5$: Two Group Setting (Advance) ]
    {
    	\pgfplotsset{
		tick label style={font=\Huge},
		label style={font=\Huge},
		legend style={font=\Large},
		}
        \begin{tikzpicture}[scale=0.4]
        \begin{axis}[
		xlabel={Time-step},
		ylabel={Error Percentage},
		grid=major,
		clip=true,
  		legend style={at={(0.5,1.2)},anchor=north, legend columns=3} ]
		\addplot table{results/budgetTestGR4/APT1_comp_subsampled.txt};
		\addplot table{results/budgetTestGR4/AugUCB1_comp_subsampled.txt};
		\addplot table{results/budgetTestGR4/UCBEM1_comp_subsampled.txt};
		\addplot table{results/budgetTestGR4/UCBEMV1_comp_subsampled.txt};
		\addplot table{results/budgetTestGR4/SR1_comp_subsampled.txt};
		\addplot table{results/budgetTestGR4/UA1_comp_subsampled.txt};
        \legend{APT,AUgUCB,UCBE,UCBEV,CSAR,UA}
      	\end{axis}
      	\label{Fig:budgetExpt5}
        \end{tikzpicture}
    }
    &
    \subfigure[0.32\textwidth][Expt-$6$: Two Group Setting (Advance) ]
    {
    	\pgfplotsset{
		tick label style={font=\Huge},
		label style={font=\Huge},
		legend style={font=\Large},
		}
        \begin{tikzpicture}[scale=0.4]
        \begin{axis}[
		xlabel={Time-step},
		ylabel={Error Percentage},
		grid=major,
		clip=true,
  		legend style={at={(0.5,1.2)},anchor=north, legend columns=2} ]
		\addplot table{results/budgetTestGR3/testUCBEMV1_0.25_comp_subsampled.txt};
		\addplot table{results/budgetTestGR4/AugUCB1_comp_subsampled.txt};
		\addplot table{results/budgetTestGR3/testUCBEMV1_256_comp_subsampled.txt};
		\addplot table{results/budgetTestGR4/UCBEMV1_comp_subsampled.txt};
        \legend{UCBEV($0.25$), AugUCB, UCBEV($256$), UCBEV($1$)}
      	\end{axis}
      	\label{Fig:budgetExpt6}
        \end{tikzpicture}
    }
    \end{tabular}
    \caption{Performances of the various TBP algorithms in terms of error percentage vs. time-step, for  six different experimental scenarios.}
    \label{fig:budgetExpt}
    \vspace{-6mm}
\end{figure}
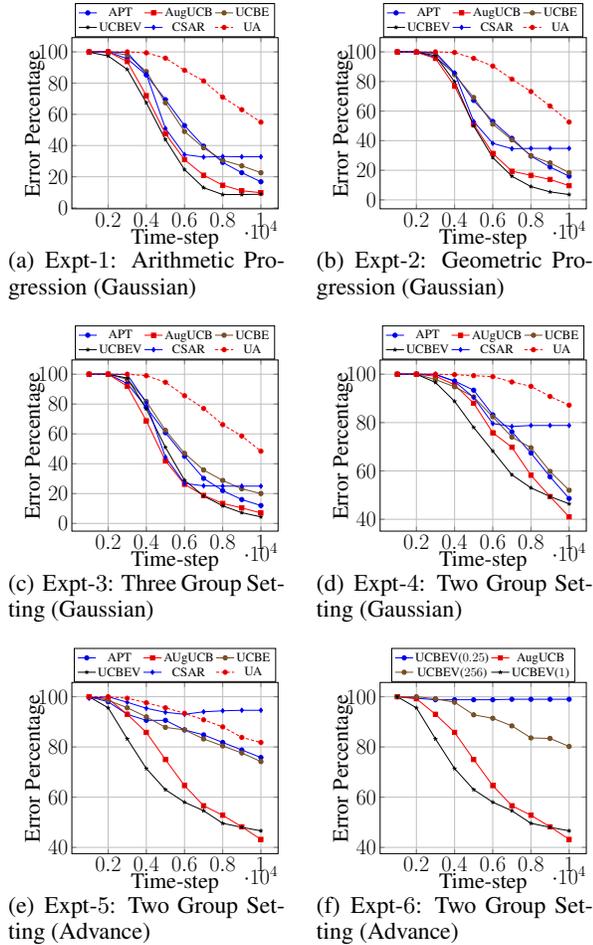

 The corresponding variances are $\sigma_{1:5}^{2}=0.5$ and $\sigma_{6:10}^{2}=0.6$, while $\sigma_{11:100}^{2}$ is chosen independently and uniform in the  interval $[0.38,0.42]$;
note that, the variances of the arms in $S_\tau$ are higher than those of the other arms.
 The corresponding  results are shown in Figure \ref{Fig:budgetExpt1},
 from where we see that UCBEV, which has access to the problem complexity while being variance-aware, outperforms all other algorithm (including UCBE which also has access to the problem complexity but does not take into account the variances of the arms).  Interestingly, the performance of our AugUCB (without requiring any complexity input) is comparable with UCBEV, while it 
 outperforms UCBE, APT and the other non variance-aware algorithms that we have considered. 	
	
\textbf{Experiment-2:} We again consider  Gaussian reward distributions. However, here the means of the first $10$ arms constitute a geometric progression.
Formally, the reward means are $r_{1:4}=0.4-(0.2)^{1:4}$, $r_{5}=0.45$, $r_{6}=0.55$, $r_{7:10}=0.6+(0.2)^{5-(1:4)}$ and $r_{11:100}=0.4$; the arm variances are as in experiment-$1$. The corresponding results are shown in Figure \ref{Fig:budgetExpt2}.  We again observe AugUCB outperforming the other algorithms, except UCBEV. 
	
\textbf{Experiment-3:} Here, the first
$10$ arms are partitioned into three groups, with all arms in a group being assigned the same mean; the reward distributions are again Gaussian. Specifically, the reward means are $r_{1:3}=0.1$, $r_{4:7}=\lbrace 0.35, 0.45, 0.55, 0.65\rbrace$ and $r_{8:10}=0.9$; as before,  $r_{11:100}=0.4$ and all the variances are as in Experiment-$1$. The results for this scenario are presented in Figure \ref{Fig:budgetExpt3}. The observations are inline with those made in the previous experiments. 


\textbf{Experiment-4:} The setting is similar to that considered in Experiment-3, but with the first $10$ arms partitioned into two groups; the respective means are $r_{1:5}=0.45$, $r_{6:10}=0.55$.
The corresponding results are shown in Figure \ref{Fig:budgetExpt4}, from where the good performance of AugUCB is again validated.

	
\textbf{Experiment-5:} This is again the two group setting involving Gaussian reward distributions. The reward means are as in Experiment-4, while the 
variances are  $\sigma_{1:5}^{2}=0.3$ and $\sigma_{6:10}^{2}=0.8$;  $\sigma_{11:100}^{2}$ are independently and uniformly chosen in the interval $[0.2,0.3]$.  The corresponding results are shown in Figure \ref{Fig:budgetExpt5}.
 We refer to this setup as \emph{Advanced} because 
here the chosen variance values are such that only  variance-aware algorithms will perform well.
Hence, we see that UCBEV performs very well in comparison with the other algorithms. However,  it is interesting to note that the performance of  AugUCB catches-up with UCBEV as the time-step increases, while significantly outperforming the other non-variance aware algorithms.

\textbf{Experiment-6:} We use the same setting as in Experiment-5, but conduct more exploration of UCBEV with different values of the exploration parameter $a$. The corresponding results are shown in Figure \ref{Fig:budgetExpt6}. As studied in \cite{locatelli2016optimal}, we implement UCBEV with $ a_{i} = 4^{i} \frac{T-2K}{H_{\sigma,1}}$ for $i = -1,0,4$. Here, $a_{0}$ corresponds to UCBEV($1$) (in Figure \ref{Fig:budgetExpt6}) which is UCBEV run with the optimal choice of $H_{\sigma ,1}$. For other choices of $a_i$ we see that UCBEV($a_i$) is significantly outperformed by AugUCB. 
	
Finally, note that in all the above experiments, the CSAR algorithm, although performs well initially, quickly exhausts its budget and saturates at a higher error percentage. This is because it pulls all arms equally in each round, with the round lengths being non-adaptive.

%
\vspace{-1mm}
\section{Conclusion}
\label{conclusion}
We proposed the AugUCB algorithm for a fixed-budget, pure-exploration TBP. Our algorithm employs both mean and variance estimates for arm elimination. This, to our knowledge is the first variance-based algorithm for the specific TBP that we have considered. We first prove an upper bound on the expected loss incurred by AugUCB. We then conduct simulation experiments to validate the performance of AugUCB. In comparison with APT, CSAR and other non variance-based algorithms, we find that the performance of AugUCB is significantly better. Further, the performance of AugUCB is comparable with UCBEV (which is also variance-based), although the latter exhibits a slightly better performance.  However, UCBEV is not implementable in practice as it requires computing problem complexity, $H_{\sigma,1}$, while AugUCB (requiring no such inputs) can be easily deployed in real-life scenarios. It would be an interesting future work to design an anytime version of the AugUCB algorithm. 

\section*{Acknowledgements} This work was supported by a funding from IIT Madras under project CSE/14-15/831/RFTP/BRAV. The work of the second author is supported by an INSPIRE Faculty Award of the Department of Science and Technology. 




\bibliographystyle{named}
\bibliography{ijcai17}


\end{document}